\documentclass[journal]{IEEEtran}

\ifCLASSINFOpdf
\else
\fi
\usepackage{url}

\usepackage{amsfonts}
\usepackage{todonotes}
\usepackage{caption, subfigure, subcaption, graphicx}

\begin{document}

\title{PyTAG: Tabletop Games for Multi-Agent Reinforcement Learning}

\author{Martin~Balla,~\IEEEmembership{Student~Member,~IEEE,}
        George~E.M.~Long,~\IEEEmembership{Student~Member,~IEEE,}
        James~Goodman,~\IEEEmembership{Student~Member,~IEEE,}
        Raluca~D.~Gaina,
        and~Diego~Perez-Liebana

        \thanks{All authors are with Queen Mary University of London.}
        \thanks{\{m.balla, g.e.m.long, james.goodman, r.d.gaina, diego.perez\}@qmul.ac.uk}
        }

\maketitle
\IEEEpeerreviewmaketitle

\begin{abstract}
Modern Tabletop Games present various interesting challenges for Multi-agent Reinforcement Learning. In this paper, we introduce PyTAG, a new framework that supports interacting with a large collection of games implemented in the Tabletop Games framework. In this work we highlight the challenges tabletop games provide, from a game-playing agent perspective, along with the opportunities they provide for future research. Additionally, we highlight the technical challenges that involve training Reinforcement Learning agents on these games. To explore the Multi-agent setting provided by PyTAG we train the popular Proximal Policy Optimisation Reinforcement Learning algorithm using self-play on a subset of games and evaluate the trained policies against some simple agents and Monte-Carlo Tree Search implemented in the Tabletop Games framework.
\end{abstract}


\IEEEpeerreviewmaketitle

\section{Introduction}

\IEEEPARstart{I}n the past decade Reinforcement Learning was shown to be a powerful paradigm, beating world champions in the board game Go~\cite{alphago} and in complex video games~\cite{alphastar, berner2019dota}. Many of these successes are on classical board games, card games and video games. Modern Tabletop Games (TTGs) are often overlooked, mainly due to the lack of a unified framework that would support their research. Previously, a few standalone modern TTGs have been implemented for research purposes, such as ``Settlers of Catan"~\cite{thomas2003real} and ``Splendor"~\cite{bravi2019rinascimento}, but all use their own interface and methods without direct compatibility with each other. To address this, the Tabletop Games (TAG)~\cite{gaina2020tag} framework was proposed to serve as a research framework with a shared interface to play and implement TTGs, with over 20 games currently in its collection. Most of the work on TAG has used Statistical Forward Planning (SFP) algorithms (e.g. Monte-Carlo Tree Search~\cite{browne2012survey}); Reinforcement Learning (RL) methods have not been employed before due to various technical challenges and the lack of support for RL algorithms.

Multi-agent Reinforcement Learning (MARL)~\cite{zhang2021multi} involves more than one agent situated in the same environment competing against each other, cooperating to achieve a common goal or a mix of the two. Advancing MARL methods has various potential real-world applications, as a lot of these problems can be naturally described as multi-agent systems: robotics, self-driving cars, video games and most tabletop games. 

The modern board games industry has been rapidly growing in the past decade with more and more games published each year~\cite{bg_market_research23}. 
Modern TTGs are designed to be played with $2$ and often more players with different types of interactions among them. These games can be competitive, where players race against each other to reach the game's winning condition first, or fully cooperative, such as Pandemic, where the players have to work together in order to win. Many games have a mix of competitive and cooperative settings; players may form a temporary alliance to stop another player from winning. 

In our previous work~\cite{balla2023pytag}, we introduced PyTAG, a python API that supports Reinforcement Learning in the collection of games that the Tabletop Games (TAG) framework offers. Our initial work explored the possibility of training Reinforcement Learning agents against two simple baseline agents: a Random agent (which samples actions with uniform probability) and the One-Step Look-Ahead agent (OSLA; using the game's forward model, it tries all the available actions and chooses the one that results in the highest evaluation score). Training against these specific opponents has served as a proof-of-concept to demonstrate the possibility of having RL agents playing these games but did not fully capture the multi-agent dynamics presented by the games. In the previous setting, the opponents could be viewed as part of the environment, which allowed training single-agent RL algorithms against them. One of the main challenges of the MARL setting is the requirement to be able to adapt to different play styles and strategies, which was not explored previously.

\begin{figure}[t]
	\begin{center}
	\includegraphics[width = .48\textwidth]{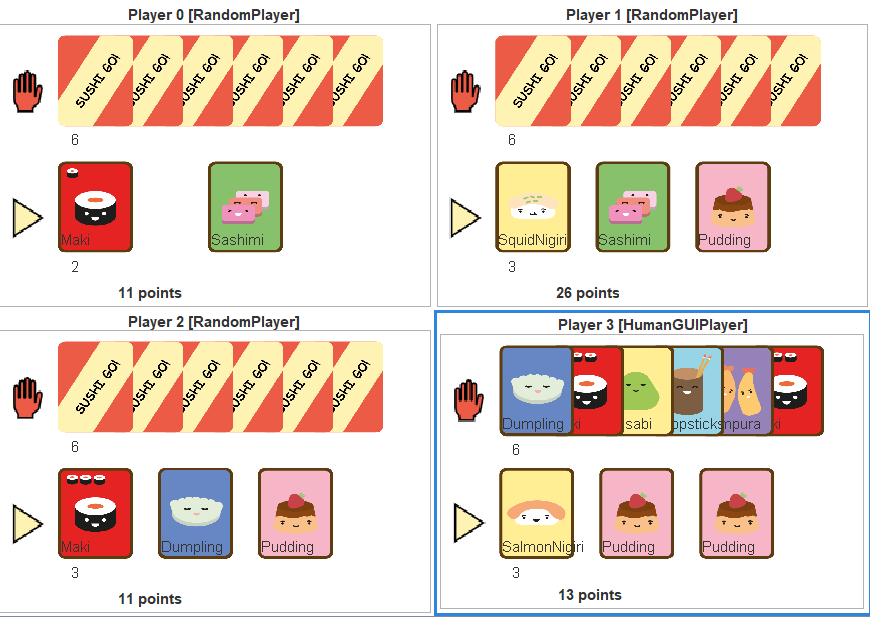}
         \caption{``Sushi Go!" Graphical User Interface in TAG}
	\label{fig:sushigo_gui}
	\end{center}
\end{figure}

To extend our previous work, in this paper we further explore the MARL setting of TTGs, by introducing a self-play setting to train Reinforcement Learning agents in PyTAG and describe the technical challenges that arise from this. By using self-play, the comparisons against the baseline agents are fairer. To evaluate the trained agents, we periodically take a snapshot of the learning agent and evaluate against the random agent, OSLA and Monte-Carlo Tree Search. In addition to self-play, we also added two new games: ``Sushi Go!" (see Figure~\ref{fig:sushigo_gui}) and ``Dots and Boxes". Sushi Go! is interesting as a strategic set collection card game in which the degree of unknown information declines over time. 
Dots and Boxes is a more complicated perfect information game than previously included, with deceptive short-term rewards.

Overall, PyTAG presents a python interface to implement Reinforcement Learning agents to play the games implemented in the Tabletop Games framework. To communication between PyTAG and TAG is done by sharing the memory locations to support running the games quickly and efficiently. TTGs come in many forms, which makes it hard to design a general representation that captures all, hence we proposed interfaces to support handling their observation and action spaces. To provide experimental results, we trained the PPO algorithm against the baseline opponents from the TAG framework to provide a proof-of-concept. In this version, we extend this work to the Multi-Agent RL setting by using self-play for training and only using the baseline agents to evaluate against. Finally, we present various challenges and opportunities that Tabletop Games and PyTAG in particular provides. 
All the code presented in this paper is available on Github.\footnote{\url{https://github.com/martinballa/PyTAG}}

\section{Background} \label{sec:background}
Reinforcement Learning (RL) studies the interaction between an agent and its environment. TTGs are typically designed for at least two players, hence we assume a decentralised MARL setting. There are multiple mathematical models applicable to such problems, but they often assume strictly turn-taking or simultaneous moves. The closest MARL setting that captures the challenges presented by TTGs is the Agent Environment Cycle (AEC) by Terry et al.~\cite{terry2021pettingzoo}. In TTGs, players often do not act simultaneously or take regular turns; to fully capture this setting we have to introduce a turn order function that manages which player needs to act in which state. 

In our MARL setting, agents are distinguished by a numerical identifier $i$, so $\pi_i$ refers to the $i^{th}$ player's policy. We assume the availability of a Turn order function $T(i | s)$ that determines the next acting player's index $i$, given the current state. The turn order is also used to randomise the initial starting order at each episode.

At each time step, the agent $i$ (specified by the Turn order function) receives an observation $o$ that contains all the observable information that the agent has access to from its point of view. The agent also receives an action mask $m$ which indicates which actions are available in the current state. Based on the observation and the action mask, the agent chooses an action $a$ and sends it to the environment. The environment updates its internal state $s$ with the agent's action and transitions into the next state $s'$. Based on the turn order function $T$, the environment gives the next observation $o'$, the next action mask $m'$ and a scalar reward $r$ to $i'$, the player who needs to make the next decision. In TTGs, players may get rewarded outside of their turn. In this MARL framework, these rewards are only awarded the next time that player needs to select an action (or at game end).

PyTAG is comparable to other Multi-agent Reinforcement Learning (MARL) benchmarks. Most other frameworks, only contain a single game with a focus on a specific aspect of MARL, such as competition~\cite{huang21microrts, whiteson2019starcraft} or cooperation~\cite{bard2020hanabi, johnson2016malmo}. PyTAG implements a shared interface to support a large collection of Tabletop Games (TTGs) which range from competitive games such as ``Exploding Kittens", to fully cooperative games such as ``Pandemic" while also including games that mix the two such as ``the Resistance". RLCard~\cite{zha2019rlcard} is a framework that supports AI research on a collection of card games. OpenSpiel~\cite{lanctot2019openspiel} is a framework that supports single and multi-agent RL with a collection of classical board and card games. Ludii~\cite{piette2019ludii} uses a description language to define board games with support for RL and planning algorithms. Unfortunately, neither RLCard, OpenSpiel or Ludii has the capacity to support the complexity that TTGs present. 

Self-play~\cite{hernandez2019generalized} has been the main driving force behind much successful multi-agent work in the past, such as AlphaGo~\cite{alphago}, OpenAI five~\cite{berner2019dota} and AlphaStar~\cite{alphastar}. Self-play trains the RL agents against previous copies of itself. This helps explore the policy space of the environment by always trying to find weaknesses of previous policies. Unfortunately, self-play can get stuck in sub-optimal strategies and is not guaranteed to converge. In various problems, a cyclic strategy loop is observed, where a set of strategies beat each other without a ``perfect" strategy that would beat all. In these cases various works attempted to converge to a Nash equilibrium~\cite{heinrich2015fictitious}. To avoid sub-optimal and cyclic strategies, AlphaStar~\cite{alphastar} proposed to use a league of different opponents with different objectives.
The Hanabi challenge~\cite{bard2020hanabi} has been a popular multi-agent benchmark in the past, posing an interesting cooperative partial-observable problem. ``Hanabi" is a card game also which is also implemented in the Tabletop Games framework. Past research on ``Hanabi" has a rich history with self-play agents and how they can be trained to remain compatible with a larger sets of partners not just the one used during training. One challenge of self-play in ``Hanabi" is that the overall objective is to train a policy that is compatible with other strategies, not just the one it was trained with.

MARL problems have been formulated under various mathematical frameworks. The two main formalisms are Extensive Form Games (EFG)~\cite{osborne1994course_efg} and Partially Observable Stochastic Games (POSG)~\cite{shapley1953stochastic}. These do not fully cover the challenges TTGs present as they assume either strictly turn-taking or simultaneous move games. PettingZoo~\cite{terry2021pettingzoo} proposed the Agent Environment Cycle (AEC), which does allow for varying turn-orders, but makes scaling difficult to multiple parallel environments. PyTAG is compatible with the PettingZoo interface, but it was not directly compatible with our implemented vectorisation and the gym style~\cite{brockman2016openai} environment wrappers, hence this was not used in our experiments. 

\section{PyTAG} \label{sec:pytag}
In our previous work~\cite{balla2023pytag}, we presented PyTAG: a python API to support Reinforcement Learning in the TAG framework. Our previous paper describes how the observation and the action spaces are handled in more detail. We give a short summary of PyTAG here, but see~\cite{balla2023pytag} for a deeper focus on the technical challenges of interfacing TAG and PyTAG, and dealing with observation and action spaces specifically.

PyTAG has been developed to support creating Reinforcement Learning agents on the collection of games available in the Tabletop Games (TAG) framework. As TTGs have a wide range of distinct characteristics, creating a standardised interface to treat the observation and action spaces for all games presents many challenges. PyTAG aims to keep the framework accessible to a wider community, hence we aim to keep the observation and action spaces flexible. We implemented interfaces in TAG to support a few selected games that could serve as a starting point to working with any of the existing (or yet to be created) games in TAG. To add support for a new game, the user is only required to write two functions: 1, a function returning a standardised observation from the game state, with flexibility on how these are represented; and 2, an action-masking function, returning which actions are available at a given state. The former allows representing observations either as a numerical vector or as a JSON string, giving control to the user's algorithm on how they want to process them for decision making. 
The action masking function is supported by TAG's engine by explicitly enumerating all available actions for each state, so to construct a mask, one only needs to write a function that matches the available actions with all possible actions.
The action masking also gives the flexibility to design more advanced action masks, such as action trees~\cite{bamford2021generalising}. 

\subsection{Games}

In this section we summarise the games we used in this work, and how we represented their actions and observations. 
\subsubsection{Tic Tac Toe}
The simplest game in TAG, ``Tic Tac Toe" is played on a $3 \times 3$ grid, with players taking turns to draw their symbol in one of the grid cells. The game finishes when a player has 3 of their symbols in a line and the player who achieves this first wins. However, if every cell is filled without a complete line of symbols, the game ends in a draw.

We represented the game as a $9$-dimensional vector: a flattened version of the board. Similarly, the action space is also a $9$-dimensional vector representing which cell the player will draw their symbol in. The action mask filters out cells which are already filled.

\subsubsection{Diamant}
A $2$ to $5$ player game, where players push their luck to gain the most treasure. Each turn, players can choose one of two actions: stay in the cave, or retreat to camp. Going back to the camp allows to safely bank your treasure. Staying in the cave allows getting more treasure, with the risk of triggering a trap and losing all your non-banked treasure. A third dummy action is to be played when the player is already at the camp, with no effect on the game; those players can only observe the other players' progress.

The observation space is a vector containing the following information from the game state: tile counters, number of gems on the last tile, and total number of gems in the cave. 

\subsubsection{Love Letter}
A $2$ to $4$ player card game, where players try to gain the most favour tokens by either being the last player standing in a round, or finishing the round with the highest valued card. Its challenges involve hidden information, and keeping track of what cards the other players may have. 

The observation space is comprised of the player's current hand (hot-encoded with 1 bit per card type), the number of each type of card in the discard pile, and how many favour tokens are held by each player. Some cards (e.g., the \textit{Guard}) have choices in their action (e.g. which player to target with the card's power once played), which leads to an increase in action size. There are only $8$ card types in the game, but due to the possible combinations of playing certain cards, there are $68$ action to choose from.
\subsubsection{Exploding Kittens}
A $2$ - $5$ player card game. Each turn, players can play as many cards as desired; however, they need to finish their turn by drawing a card. If an \textit{Exploding Kitten} card is drawn, the player will lose the game, unless they have a defensive card available. The deck contains $n - 1$ \textit{Exploding Kitten} cards, where $n$ is the number of players. The game has a reactive turn order: some cards can be used at any point, others allow for cards to be taken from other players. These challenges make the game interesting from an RL perspective.

In terms of the observation space, there are counters for each card type contained in the player's hands. In addition, players observe the number of cards in each opponent's hand, how many cards are in the draw pile, and, finally, which game phase is in play. The game phase informs the player whether it needs to take actions as normal or if it needs to react to some event. The action space contains drawing a card, playing a card from hand or reacting to an event resulting in $43$ possible actions.

\subsubsection{Stratego}
A $2$-player strategy game, it takes place on a $10$ x $10$ grid. A player can win by capturing the opponent's flag, or if the opponent is unable to make any moves. The game's main challenge is hidden information: the enemy's unit types are unknown until the unit is involved in combat.

The observation space is a representation of the $10$ x $10$ grid, with unit types being hot-encoded. $27$ feature maps represent each player-exclusive unit type. The action space for the game is quite large: $4400$ actions. This is due to each player having $40$ units. In addition, the \textit{Scout} unit type has the ability to move to any non-diagonal tile, as long as it is unoccupied.

\subsubsection{Sushi Go!}
``Sushi Go!" (shown in figure~\ref{fig:sushigo}) is a card game a $2$ - $5$ player drafting card game where players aim to accumulate points by playing cards, with points being updated at the end of each round. At the end of each turn, players pass their hands to the next player in a clockwise rotation. The main challenge of this game is the hidden information: in addition to the players not being able to see opponents' hands, not all cards are in play at any one time.

The observation space consists of an encoding of all the cards visible from the player's point of view, so this includes the player's current hand and all the played cards displayed on the table. Additionally, it includes scoring-related information. The action space represents which card the player will play from its hand, or use a previously played \textit{Chopstick} card (which allows playing two cards from the same hand at once) resulting in $20$ possible actions.

\subsubsection{Dots And Boxes}
``Dots and Boxes" is a game for $2+$ players, where players aim to complete as many boxes as possible on the grid the game takes place on. The grid is made up of dots, which players can join together (horizontally or vertically) to create edges between contiguous dots. If a player draws an edge which completes a box, they gain a point and can draw again. This combo factor comprises the main challenge of this game: long-term planning is necessary to achieve these combos, as well as balancing between short and long-term rewards.
The observation space is represented as the $82$ edges, with each edge having a value of 1 if the player drew it, a value of 0 if missing, and a value of -1 if it was drawn by another player. There are $82$ possible actions, one for drawing each edge, but as the game progresses the valid actions are reduced by one at each step.

\subsection{Challenges of Tabletop Games}

The interaction between players in TTGs presents various MARL settings: fully competitive, cooperative or a combination of both. As presented in Section~\ref{sec:background}, in the general case TTGs can have arbitrary turn orders, but there are many strictly turn-taking and also simultaneous move games that can be modelled using other MARL formulations. Additionally, in our case, we assume a decentralised setting, but many games could make use of a centralised controller - for instance in the game Pandemic up to $4$ players need to cooperate with a common goal, while in some games a player may have multiple units to control on the board (for instance the overlord in Descent, or Dungeon and Dragons in general). 

TTGs often have a large amount of hidden information that comes in many forms, such as opponent's hands and draw decks. The result of the opponents' actions is often not visible until the end of the game; for instance, in Exploding Kittens, players often collect cards in their hand, which are drawn face down. Therefore, other players have no knowledge about the cards the opponents have until those cards are played. 
TTGs also have a high level of stochasticity, by drawing shuffled cards, rolling dice and randomised initial setups to support replayability for human players. Some games have multiple winning and losing conditions, and deceptive scoring systems~\cite{anderson2018deceptive} where maximising immediate rewards can lead to suboptimal strategies (i.e: Push your luck games such as Diamant, or in Catan where it is better to build an `engine' to acquire the specific resources needed in the late-game and not necessarily go for immediate points now). Additionally, games without a scoring function often have very sparse rewards that require good exploration strategies to find viable policies. In TTGs, players often receive points during the opponent's turn or scores may only get revealed at the end of the game, which poses an interesting credit assignment problem. 

\subsection{Reward Functions}

Good reward functions are essential for Reinforcement Learning algorithms, as they define the agent's learning objective. Tabletop games have many kinds of scoring functions. Most commonly games have a clearly visible individual score, such as Settlers of Catan where the players need to collect $10$ points. Other games have specific winning conditions such as Stratego, where the player needs to capture the opponent's flag. There are also games like Exploding Kittens where players are eliminated throughout the game and the last standing player is the winner. For RL, these objectives may be very sparse or even misleading (short-term vs long-term rewards). TAG allows implementing game-specific heuristics $h: S \rightarrow \mathbb{R}$, which allows defining custom reward functions for each game, but it requires some engineering effort to do so. In the context of TAG, Goodman et al.~\cite{goodman2023followtheleader} proposed a set of game-agnostic reward functions to support Statistical Forward Planning algorithms to play at a higher level. We made these reward functions available for agents implemented in PyTAG as an attempt to lower the entry barrier and to support developing new algorithms in the framework. With this addition, PyTAG supports the following game-agnostic reward functions:

\begin{itemize}
    \item Terminal (Default); gives the winning player $+1$, the loser $-1$ and in case of a draw $0.5$ reward. 
    \item Score; The current player's in-game score
    \item Leader;  - The difference in score between the player with the highest score and the current player's score
    \item Ordinal; - The current player's relative position in the game ranking
\end{itemize}
The reward function involving scoring (Score and Leader) are only available in games that have a measurable score, while the others are available for all games. 

\section{Method}

In our previous work, we presented agents that were trained directly against specific opponents (Random and OSLA). Training against stronger agents, such as MCTS slows down the training significantly as SFP methods require time for planning for each action selection. 
Training against specific opponents also means RL agents may just learn to exploit the opponents without learning to play the game well more generally. Training against a specific opponent reduces the MARL setting into a single-agent setting as during training the opponents can be modelled as part of the environment. In this work, we explore the additional challenges that come with training in the MARL setting by implementing a self-play setting. In this, the RL agent is only trained against previous versions of itself with periodic evaluations against the baseline agents in order to measure relative performance against them.

\subsection{Technical Challenges of Self-play}

\begin{figure}[t]
	\begin{center}
	\includegraphics[width = .45\textwidth]{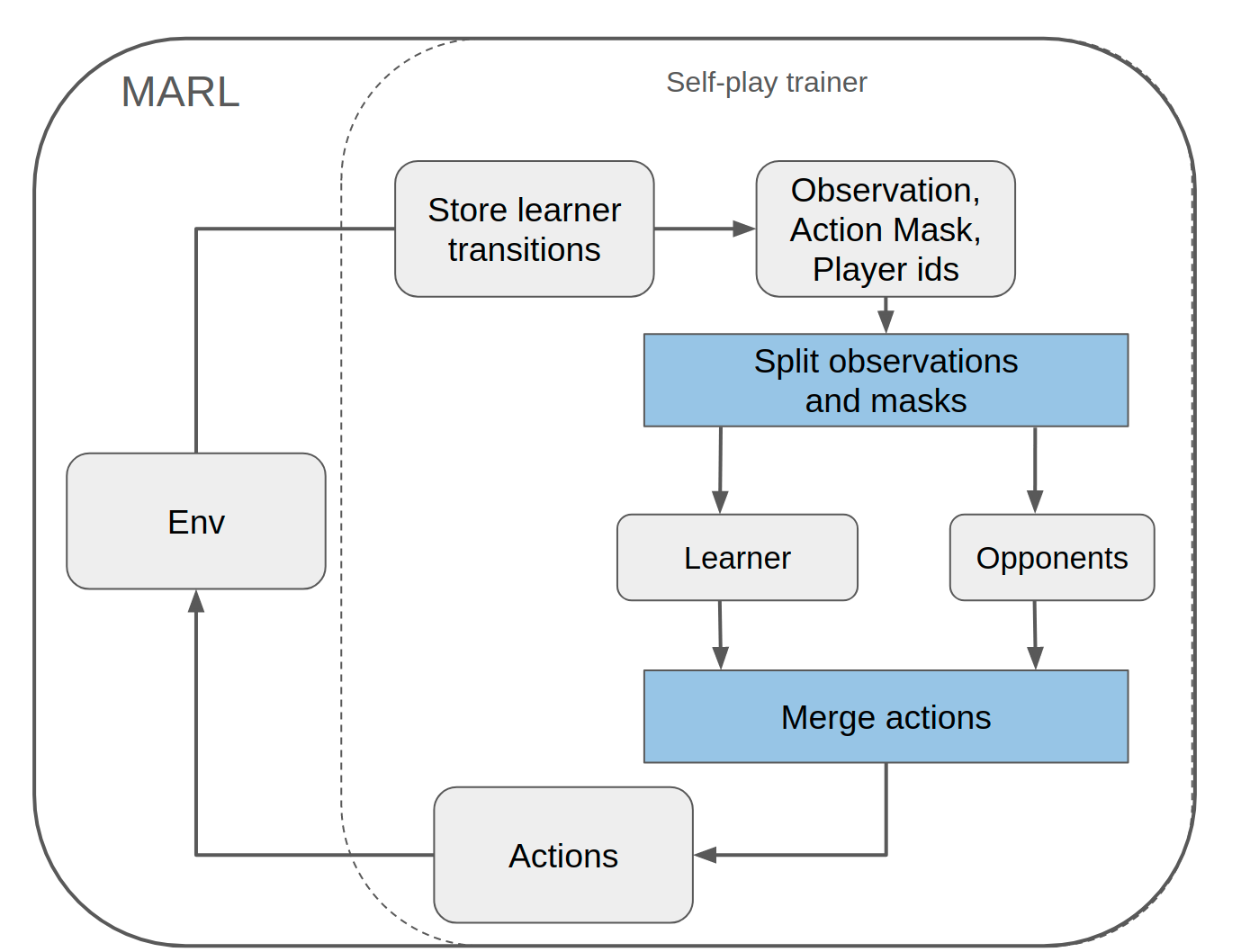}
         \caption{The overall self-play setting used in our experiments.}
	\label{fig:marl_loop}
	\end{center}
\end{figure}

Working in a full MARL setting in tabletop games poses new challenges compared to previous works. In the previous iteration of PyTAG, training was against the baseline agents in TAG, so the RL agent was only asked to make decisions when it was its turn. In contrast, in the MARL setting, we need to manage all agents' action selections. Other similar frameworks assume either a strictly turn-taking setting or simultaneous move action selection. Most tabletop games do not fall into either of these categories, so we have to assume that at each step an arbitrary agent may need to make a decision. Our proposed training setup is shown in Figure~\ref{fig:marl_loop}. 

In this work, we use self-play to train MARL policies. Throughout training with self-play, the current version of the learner is regularly saved and added to a pool of checkpoints which are subsequently re-used as opponents. To stabilise training the sampled checkpoints are kept as opponents for an extended period. As previously highlighted, in TTGs the initial player often has an advantage or disadvantage, or some asymmetry. For self-play and later for evaluation it is crucial to train agents that can play well from any player position. With the episode resets, a random player position is sampled to take the role of the learner player. At each time step during training the observations and masks are split into two groups, those that belong to the learner agents and those for the opponents. Depending on whether the learner or an opponent is required to take an action, the observations with the action masks are given to the corresponding agents. After the actions are selected, they are merged back together, maintaining the original order, so each environment gets the corresponding action. After this, the environment is updated with the actions and the cycle repeats for the next steps. After taking steps, the learner's transitions are stored in a buffer to be used for training later.   

Normally, using RL with vectorised environments, the learner takes the same number of steps in all environments, resulting in a regular-shaped matrix that can be used for optimisation. With varying turn-orders, this is often not the case, as some players may take more actions than others. To tackle this, instead of waiting for all agents' transitions to fill up a predefined length, we do an update as soon as one player in the environment reaches the desired length. The number of transitions used for optimisation varies, however, we make use of every single transition. Waiting for all the buffers to fill up would lead to new transitions coming in for some already full buffers, which may result in discarding the extra transitions, and not using them for optimisation.

To avoid converging to suboptimal policies with self-play we used various hyper-parameters to control the matchups during training. First, we specify how many checkpoints we keep in the opponent pool. Additionally, we define how regularly checkpoints are added to the training pool. Finally, the opponent policy should be replaced with a certain frequency. Keeping opponent policies for a longer period may result in one-sided matches where the learning agent is not challenged enough; while keeping only the most recent policies might result in converging to specific strategies that may not generalise to other opponents. We also added a hyper-parameter to bias the opponent sampling probability to take the latest checkpoint, otherwise with uniform sampling the new checkpoints get sampled less frequently overall. 

\begin{figure*}
	\begin{center}
        \includegraphics[width = .24\textwidth]{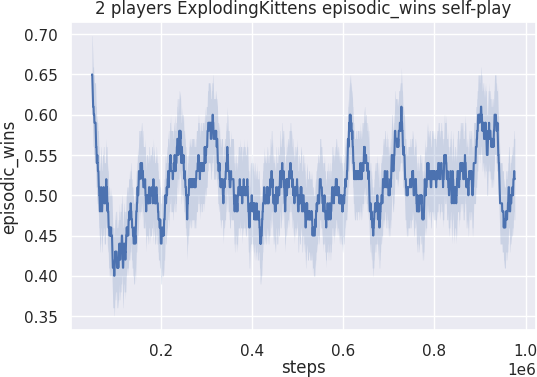}
        \includegraphics[width = .24\textwidth]{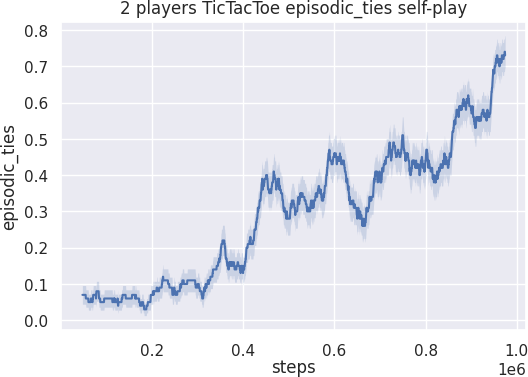}
        \includegraphics[width = .24\textwidth]{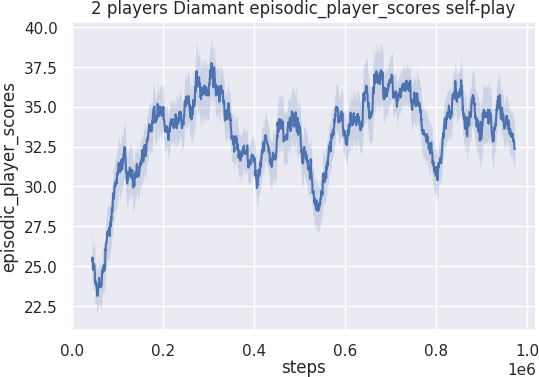}
        \includegraphics[width = .24\textwidth]{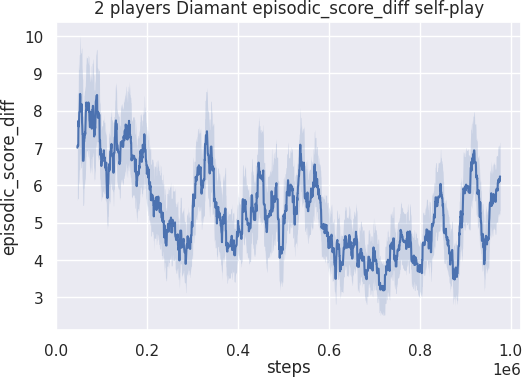}
	\caption{Some metrics to highlight the outcomes observed during self-play. All plots show the running mean of the metric from the last 100 episodes with the standard error shown with the shaded area. From left to right: Exploding Kittens win rate, Tic Tac Toe tie rate, learner's achieved scores and score differences (score of the winner minus learner's score) in Diamant.}
	\label{fig:selfplay_insights}
	\end{center}
\end{figure*}

\begin{figure}
	\begin{center}
	\includegraphics[width = .24\textwidth]{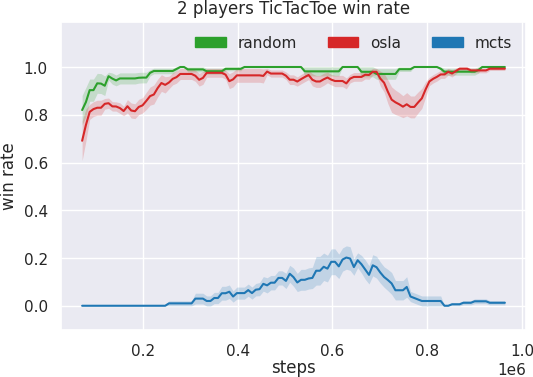}
	\includegraphics[width = .24\textwidth]{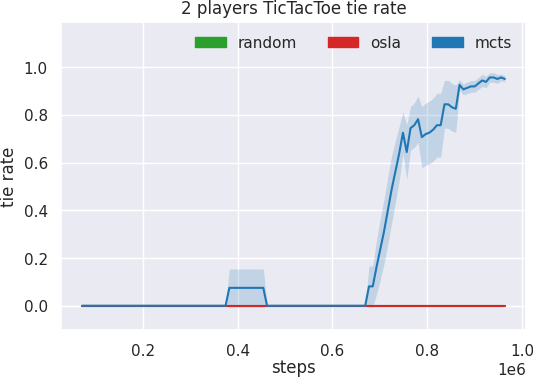}
	\caption{``Tic Tac Toe" evaluation win and tie rate.}
	\label{fig:ttt}
	\end{center}
\end{figure}

\begin{figure}
	\begin{center}
	\includegraphics[width = .24\textwidth]{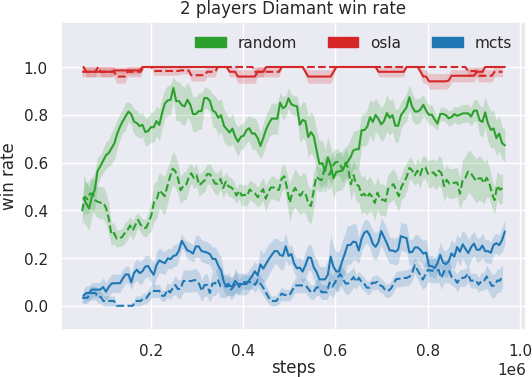}
	\includegraphics[width = .24\textwidth]{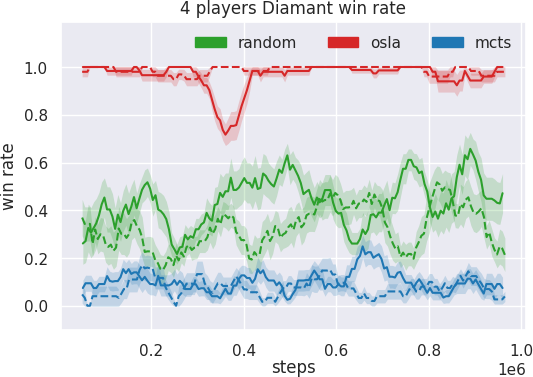}
	\caption{Evaluation win rate against the baseline agents in 2 and 4-player ``Diamant". Solid lines show agents trained with only the Terminal reward. Dashed lines used the score as reward.}
	\label{fig:diamant}
	\end{center}
\end{figure}

\subsection{Actions and Observations}
Compared to Video Games, Tabletop Games have fewer restrictions in terms of actions and observations. Creating a game-agnostic general representation is a challenging task. Instead, PyTAG uses game-specific extractors to create representations of the action and observation spaces. To add support to a new game, users only need to implement two interfaces: 1, to extract observations from the game states (either as a vector or as a JSON representation) and 2, a function that creates an ordered list of actions. Currently, $8$ games are supported and adding support for new games is straightforward.

Actions in tabletop games can be more abstract than ones in video games. This is due to them not being restricted by an input device such as a controller. In complex tabletop games, the action space can be very large, and often combinatorial. This complexity is further compounded by aspects such as multi-action turns, reactions, and extended actions (actions which have multiple sub-actions).

At each time step TAG computes the available legal actions for the current state by default. PyTAG makes use of the list of available actions to create an action mask, an indicator vector representing the available actions with 1s and the unavailable ones with 0s. With the action masks, we can assure that the RL agents can only pick from the valid actions. 

\section{Experimental Setup}

\begin{figure*}[!t]
	\begin{center}
	\includegraphics[width = .24\textwidth]{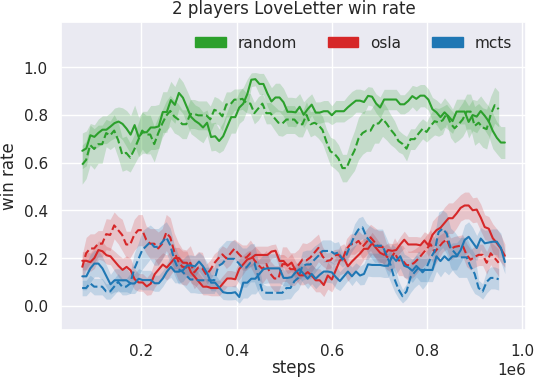}
	\includegraphics[width = .24\textwidth]{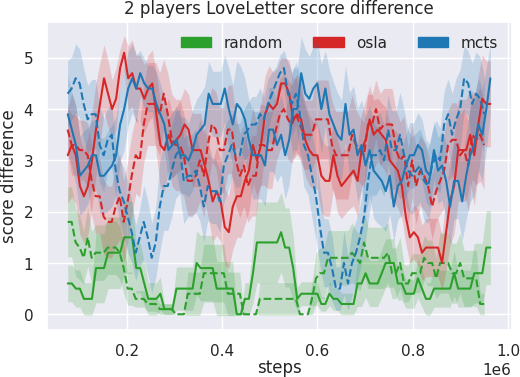}
        \includegraphics[width = .24\textwidth]{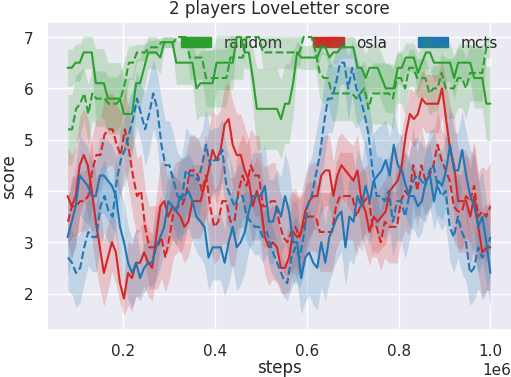}
	\caption{2-player ``Love Letter" evaluation results. In a two-player game, the player who collects $7$ points wins the game. Shown from left to right: win rate, score difference between winner and learner, and finally the learner's final score. }
	\label{fig:loveletter}
	\end{center}
\end{figure*}

In our self-play experiments, we follow the logic shown in Figure~\ref{fig:marl_loop}. For most games, we used simple fully-connected Neural Network architectures as presented in our previous work~\cite{balla2023pytag}. For Stratego, we used a convolutional neural network due to its grid-shaped board. Our self-play implementation extends the PPO implementation from CleanRL~\cite{huang2022cleanrl}, which was also used in our previous experiments. 

We list the final parameters used in all the experiments presented in this paper in this section.  TicTacToe and Diamant were used to tune the hyper-parameters. All agents were trained for $1e6$ interactions with the environment (same budget as used previously) across $3$ seeds. We ran experiments for more time steps ($2.5e6$) but we did not observe significant changes in most games, hence the previous budget was maintained for easier comparisons. Learning rates in the range $[0.01, 0.0001]$ were tried, and $0.001$ was found to work best. For self-play, the pool of opponents stores the last $10$ checkpoints, and new checkpoints are added every $100,000$ interactions with the environment. A new opponent is sampled every $20,000$ steps from the pool of opponents with $50$\% probability to sample the latest checkpoint. 

When training using self-play, it is essential to periodically evaluate the policies to monitor learning progress. In our experiments, we evaluated the latest policy against the baseline agents implemented in TAG every $20,000$ steps, $5$ episodes per opponent. The evaluation episodes are only used to give a relative estimate of the RL agent's performance, but not for training. Compared to our previous work, this setting is a fairer comparison against the baseline agents, as the RL agents are not trained to beat them specifically. Due to the time required for MCTS to make decisions, our previous work was limited to training against Random and OSLA. This setup allows comparisons against MCTS without unreasonable overhead. 

One of our main objectives of PyTAG is to provide a research benchmark that is suitable for running experiments quickly. Hence we provide all the code required to vectorise the games, collect statistics and train the RL agents used in this work. All the experiments were run on a machine with $16$ CPU cores without making use of a GPU. With self-play, we introduce an additional overhead by using multiple neural networks for inference during training, but even in this setting an agent can be trained in a few hours on a normal laptop. 

\subsection{Opponents}
This section describes the opponents (all part of the TAG framework) that used to evaluate the RL agents. The simplest agent is \textit{Random}, which samples a valid action at each step from the list of available actions with uniform probability. \textit{One Step Look Ahead} (OSLA) is another simple agent that uses a forward model to try all available actions at each step. The action that leads it to the next state with the highest score is chosen. In case of a tie, an action is chosen with uniform probability among the tied actions.

Finally, \textit{Monte Carlo Tree Search} (MCTS) is also used as an opponent to evaluate the RL agents. MCTS is a Statistical Forward Planning method that uses the game's built-in forward model to simulate rollouts. MCTS iteratively builds a tree with estimated state-action values and when a time budget is reached it returns the best available action. MCTS is a well-known algorithm for general video and board gaming with various extensions proposed by the community to tackle certain problems. In this work, we use the default version of MCTS implemented in the TAG framework. This version of MCTS is not tuned for any game specifically but gives a reliable comparison to the relative strength of the RL agents.

\section{Results}

\begin{figure}[!t]
	\begin{center}
	\includegraphics[width = .24\textwidth]{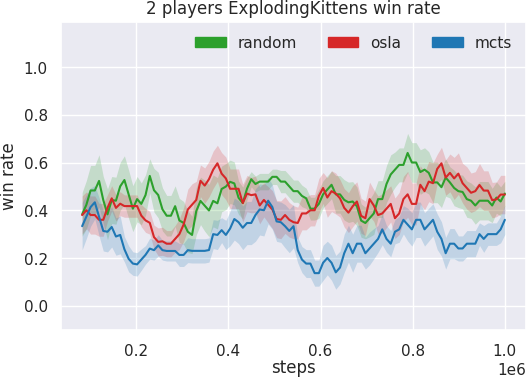}
        \includegraphics[width = .24\textwidth]{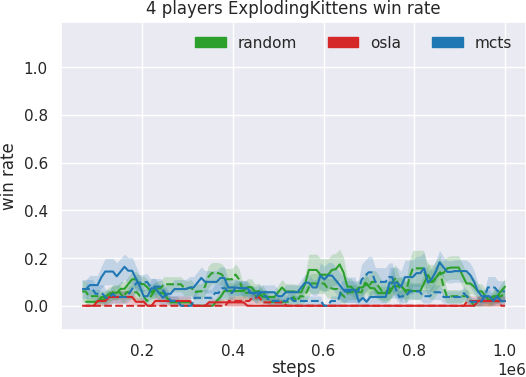}
	\caption{2 and 4-player ``Exploding Kittens" win rates.}
	\label{fig:ek}
	\end{center}
\end{figure}

\subsection{Self-play results}
Training RL agents using self-play requires further evaluations as playing against previous copies of itself often leads to close games. Just looking at win-rates during self-play gives little insight of how the agents are improving due to most games result in a close to $50\%$ win-rate, hence in figure~\ref{fig:selfplay_insights} we show multiple metrics for the chosen games to highlight how the agents improve throughout training. For clarity, we picked only the two-player runs to reduce the noise shown in the four-player runs. As we can see in Exploding Kittens, the win rate remains close to a 50\% which we observed with most games. In some games the outcome is binary, the player either wins or loses, but we also have games where fairly skilled players can tie games. As an example, in TicTacToe the win rate drops to near zero with agents getting better at playing the game. In many games, there is no scoring but rather specific win condition(s) that end the game when they are met. On the other hand, games with a clear score give a good indication of how agents get better at playing the game. For example, in Diamant, we can see that agents get higher scores as the learning progresses. In addition, we can observe the score difference between the winner and the learner which shows that the games are getting closer, which indicates that the learner is getting challenged at the right level.

\subsection{Evaluation results}

\begin{figure}[!t]
	\begin{center}
	\includegraphics[width = .24\textwidth]{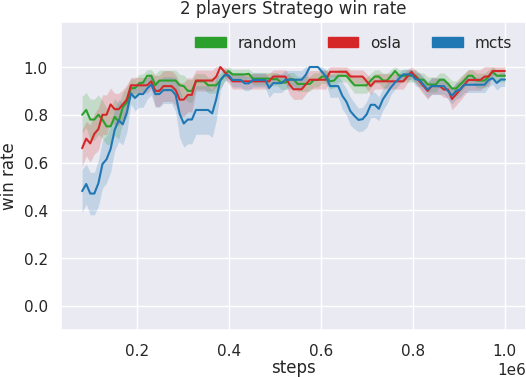}
        \includegraphics[width = .24\textwidth]{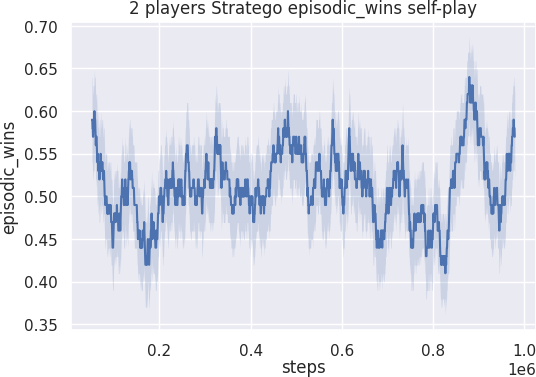}
	\caption{``Stratego" wins vs baselines (left) and  self-play (right).}
	\label{fig:stratego}
	\end{center}
\end{figure}

\begin{figure}
	\begin{center}
	\includegraphics[width = .24\textwidth]{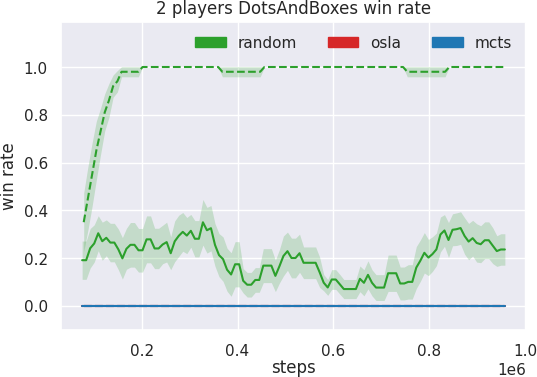}
	\includegraphics[width = .24\textwidth]{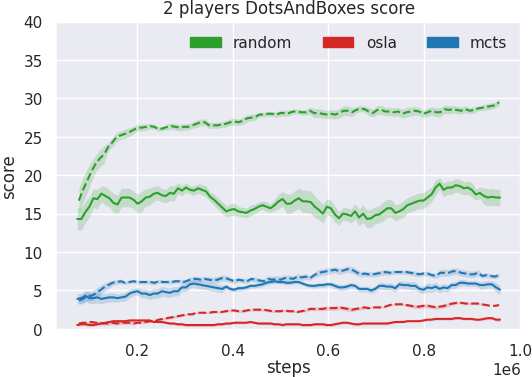}
	\caption{Evaluation win rate and scores in ``Dots and Boxes".}
	\label{fig:dandb}
	\end{center}
\end{figure}

As discussed above, just using self-play makes it hard to evaluate how well the trained agents can play. To get a better estimate of their performance we used regular evaluations against the other agents in the framework. We tracked various metrics against each opponent, such as the RL agent's win rate, tie rate, loss rate and score differences from the winner. Note that all metrics refer to the RL agent's point of view, i.e: win rate refers to how often the RL agent wins. 

Figure~\ref{fig:ttt} shows the evaluation results on TicTacToe, where the RL agent has quickly learned how to beat the simple players Random and OSLA. Although MCTS is a fairly strong opponent in TicTacToe, the RL agent learns to beat it around 20\% of the time towards midway of the training. However, it loses in most of the other cases playing with a risky strategy. Towards the end, PPO converged to playing a more conservative strategy, tying around 90\% of the games against MCTS. 

Diamant is a push-your-luck game where the players need to risk their potential score to score higher. Figure~\ref{fig:diamant} shows the win rates against the baseline agents. Looking back at the self-play results (figure~\ref{fig:selfplay_insights}), on Diamant we can see the RL agents learned to risk more and explore deeper into the cave. 

The results on Love Letter are shown in figure~\ref{fig:loveletter}. Love Letter was also a challenging game to master for the RL agent, as it requires memorisation and inference about the unknown opponent cards. In Love Letter the RL agent quickly learns how to play well against the random player, but fails to converge to a strong policy during training. Looking at the score differences, even against Random the RL agent plays close games throughout training while it often gets beaten by the other agents by a large margin (3-4 points).

Exploding Kittens (figure~\ref{fig:ek}) is a highly stochastic game with a major element of luck. This stochasticity is reflected in the learning results where the RL agent has similar performance to OSLA and Random, and wins against MCTS around 30\% of the time without showing any visible improvement over the course of training. Interestingly, on the four-player setting we observe a significant drop in performance which highlights how having more than $2$ players affects the learning dynamics. 

Stratego is a highly complex game with a large amount of hidden information and a large action space. Figure~\ref{fig:stratego} shows that during self-play the win rates deviate a lot, which we hypothesise to be due to the agent exploring the strategic depth of the game. In contrast to noisy self-play win-rates against the other players, PPO quickly finds a policy that consistently beats the baseline agents. We believe that the strong performance on Stratego might be due to other players struggling more with higher action space sizes and partial observability.

The first new game added to PyTAG is ``Dots And Boxes". The results against the baseline agents on Dots and Boxes are shown on figure~\ref{fig:dandb}. One shortcoming of our setting is that the observations in our experiments were encoded into a vector representing the owner for each edge without the spatial information on how they are connected. In addition to that both OSLA and MCTS were good at preventing the RL player from scoring which leads to a difficult training setting. Even for human players Dots and Boxes can be challenging as winning can be challenging against expert players that require calculating many turns ahead, and involve sacrificing points to be able to play in the right turn-order~\cite{barker2012solving}. Due to these challenges, the RL agent struggled to learn a good policy and lost all the games against OSLA and MCTS but managed to beat the Random player all the time when using the Score as reward. When it only used the terminal rewards it performed worse than random due to the combination of the large action space, lack of spatial information and sparse rewards. Even in this setting, we observe a clear sign of improvement as PPO quickly learns to beat Random and its score gradually increases over the course of training. 

\begin{figure*}
	\begin{center}
        \includegraphics[width = .24\textwidth]{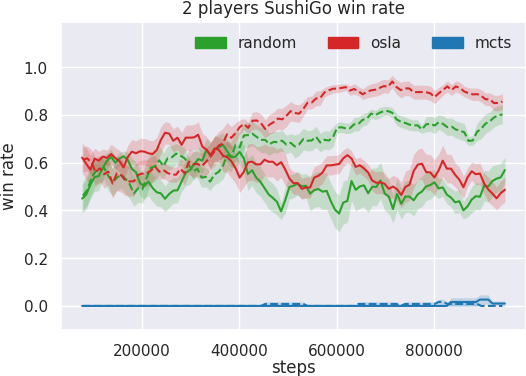}
	\includegraphics[width = .24\textwidth]{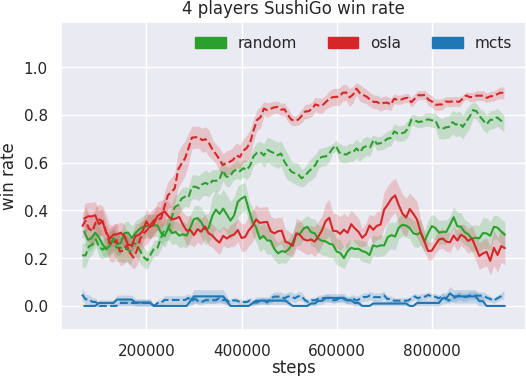}
        \includegraphics[width = .24\textwidth]{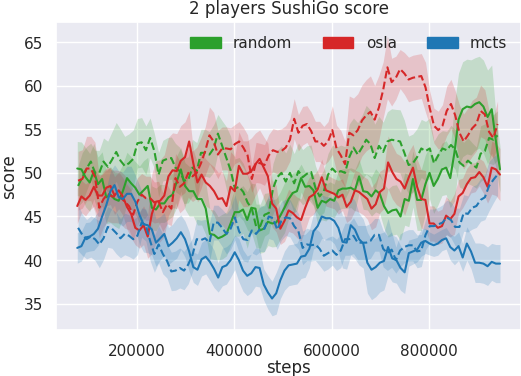}
        \includegraphics[width = .24\textwidth]{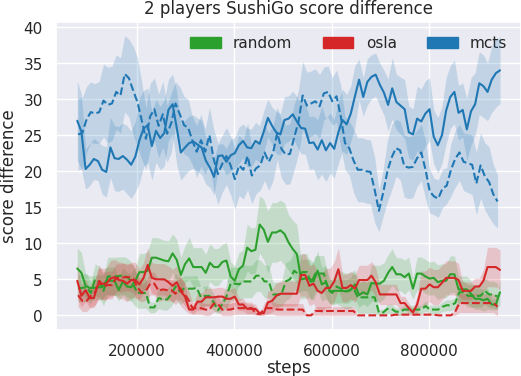}
	\caption{``Sushi Go!" evaluation results. From left to right; 2 and 4-player win rates, episodic player scores and score differences between winner and learner agents. Solid lines show training with only Terminal reward. Dashed lines use score as reward.}
	\label{fig:sushigo}
	\end{center}
\end{figure*}

The next addition to the collection of games supported by PyTAG is ``Sushi Go!". 
This game has an interesting mechanic where each player gets a hand of cards among which they simultaneously play a chosen card and then give their hand to the next player. Initially, Sushi Go! contains a large amount of hidden information which gets revealed completely as the hands are exchanged. Sushi Go! is a good challenge to test the memory capabilities of RL agents. The baseline agents in TAG know what the revealed cards are, so they get better information for decision-making, but this information is not represented explicitly in the observations given to the RL agents. Figure~\ref{fig:sushigo} presents the evaluation results obtained in our experiments. The dashed lines represent cases where PPO was trained using the game score as reward, while the solid lines only used the game outcome (win/loss) as reward. Against Random and OSLA, PPO learns a good policy in both 2 and 4-player setups. Against MCTS, the RL agents do not perform that well, but that may be due to MCTS having an effectively perfect memory as the history is encoded within the full state, while the RL agents only make decisions based on their current observation which does not include history. Looking at the score differences on figure~\ref{fig:sushigo} we can see that as the training progresses, the RL agent loses by fewer points, so with more training PPO may become more competitive. Using the score as reward has resulted in a significantly better performance against the simpler agents both in win rate and accumulated score at the end of the game.

Table~\ref{tab:results} presents the results obtained by evaluating against the baseline agents: Random, OSLA and MCTS. To facilitate comparisons, we added the results from our previous paper in Table~\ref{tab:cog_results}. Note that the performances shown in the table from the previous paper only show the cases when the agent was evaluated against the same opponents as it was trained against. Making direct comparisons against the results from the previous work is unfair, as the agents are not expected to generalise beyond the setting they were trained on. However, it can serve as a comparison point to show what can be achieved if the RL agent is trained particularly against that opponent. 

In most games where the score is available, it leads to better performance. ``Diamant" is a game where you have to take risks to gain more points, using the score as reward has resulted in worse performance on both $2$ and $4$ player settings.

One challenge with self-play is that agents need to train against the right opponent: if the opponent is too weak, the training becomes less ``interesting", as it can just easily beat that version. On the other hand, only using the latest checkpoints can lead to sub-optimal policies. One example to highlight is ``Tic Tac Toe", where training against Random and OSLA both results in high win-rates, but those policies do not transfer when evaluated against MCTS. Using self-play, the agent is able to play well against any of the baselines.  

Overall, using self-play we can train more robust agents that can play against multiple opponents. Unfortunately, even with self-play, agents are not guaranteed to converge. Many games present cyclic strategies, hidden information and high stochasticity in outcomes that may result in never reaching a 100\% win rate. Further training and game-specific information could significantly improve the results presented in this paper, leaving plenty of opportunities to study TTGs in PyTAG. 

\begin{table}[!t]
\begin{center}
\begin{tabular}{|c|l|l|r|r}
\hline
\textbf{Players} & \textbf{Game}   & \textbf{Random} & \textbf{OSLA}  \\
\hline  
         &Tic Tac Toe &          0.96 (0.02) &  1.0 (0.00)   \\
         &Diamant &            0.82 (0.04) & 0.85 (0.04)  \\
    2    &Love Letter &          0.93 (0.03) & 0.52 (0.05)  \\
         &Exploding Kittens &     0.74 (0.04) & 0.77 (0.04) \\
         &Stratego &              0.26 (0.05) & 0.02 (0.01) \\
\hline         
          &Diamant &          0.86 (0.03) & 0.75 (0.04)   \\
    4     &Love Letter &         0.59 (0.05) & 0.21 (0.04) \\
\hline    
\end{tabular}
\caption{Results from our previous work~\cite{balla2023pytag}. All agents were trained against the specific opponent they were evaluated against. As scoring function, the Terminal (Default) reward function was used. The win rates shown in the table are the average win rates on the last $100$ episodes during training.}

\label{tab:cog_results}
\end{center}
\end{table}

\section{PyTAG Opportunities}
With the recent success of Large Language Models (LLMs)~\cite{brown2020language_gpt3} there are more and more works that take a game environment and use such models for game playing~\cite{wang2023voyager, klissarov2023motif}. Tabletop games typically have a large amount of textual information that was difficult to handle in the past. For instance, the game ``Dominion'' has a large number of distinct cards with lots of textual information on the cards. To play such games the player first needs to get a good understanding of the game state, hence it would need to learn first how to interpret the cards. Using LLMs, these cards could be interpreted right away without careful engineering. The flexible interfaces provided by PyTAG allow the implementation of custom text-based state extractors.

\begin{table}[!t]
\begin{center}
\resizebox{\columnwidth}{!}{%
\begin{tabular}{|c|l|l|r|r|r|r|r|}
\hline
\textbf{NP} & \textbf{Game} & \textbf{Rewards} & \textbf{Random}  &   \textbf{OSLA} &  \textbf{MCTS} \\
\hline  
         &Tic Tac Toe & Terminal &            1.0 (0.02) &     1.00 (0.00) &    0.05 (0.06) \\
         &Diamant & Terminal &               0.72 (0.21) &    0.99 (0.06) &  0.26 (0.18) \\
         &Diamant & Score &                0.56 (0.21) &    0.99 (0.05) &  0.13 (0.13)  \\
         &Love Letter & Terminal &            0.8 (0.17) & 0.22 (0.19) & 0.17 (0.16) \\
         &Love Letter & Score &            0.75 (0.19) & 0.22 (0.19) & 0.16 (0.17) \\
    2    &Exp. Kittens & Terminal &        0.51 (0.23) & 0.49 (0.18) & 0.31 (0.20)  \\
         &Stratego & Terminal &               0.91 (0.12) & 0.95 (0.10) & 0.88 (0.16)  \\
         &Dots and Boxes & Terminal &      0.23 (0.22) & 0.0 (0.00) & 0.0 (0.00)  \\
         &Dots and Boxes & Score &          1.0 (0.00) & 0.0 (0.00) & 0.0 (0.00)  \\
         &Sushi Go! & Terminal &           0.53 (0.18) & 0.64 (0.16) &  0.0 (0.00)  \\
         &Sushi Go! & Score &              0.81 (0.11) & 0.88 (0.08) & 0.01 (0.02)  \\
\hline         
         &Diamant & Terminal &                 0.5 (0.20) &    0.98 (0.05) &  0.13 (0.13) \\
         &Diamant & Score &                 0.33 (0.19) &   0.99 (0.04) &  0.07 (0.11) \\
         &Love Letter & Terminal &             0.28 (0.19) &    0.03 (0.07) &   0.01 (0.04) \\
    4    &Love Letter & Score &            0.42 (0.20) &    0.09 (0.13) &   0.03 (0.07) \\
         &Sushi Go! & Terminal &    0.27 (0.14) &    0.31 (0.15) &   0.13 (0.13) \\
         &Sushi Go! & Score &      0.76 (0.11) &    0.88 (0.08) &   0.05 (0.04) \\

\hline    
\end{tabular}
}%
\caption{Evaluation win rates against the baseline agents implemented in the TAG framework. NP stands for number of players. All agents were trained using self-play without training against specific opponents. Win rates are averaged across the last $20$ evaluations ($5$ episodes for each, totalling $100$ evaluations), equivalent to the last 20\% of the training.}
\label{tab:results}
\end{center}
\end{table}

Another case where LLMs could be beneficial is on games requiring interaction among players using natural language. For instance, in ``Settlers of Catan'', instead of choosing from a list of pre-defined trade offers, these agents could propose their own trades, including reasoning on why those would benefit other players. Some TTGs are designed to be played by larger groups of players that involve free communication. Social deduction games is a popular genre where players are given a hidden role with a personalised or team-based objective. Discussion game phases are common where players can communicate, aiming to reach common decisions and deduce each other's roles. AIwolf~\cite{toriumi2017aiwolf} is a competition that focuses on the game Werewolf, with various tracks involving a predefined protocol for communication or natural language, assumed to be compatible with human players. TAG already implements the ``Resistance'', which is another popular social deduction game that, like AIwolf, includes a phase where players discuss lines of actions (i.e., team formation). Another recent combination of LLMs with board games is CICERO~\cite{meta22diplomacy}, used to play the game ``Diplomacy'' where players are given time to discuss their intentions between turns in natural language. 

In this work, we explored self-play to train more robust agents, as opposed to training against specific opponents. However, this is only a stepping stone. In order to better explore the strategic depths of TTGs, others proposed training a league of agents with different objectives~\cite{alphastar} or using a population of diverging agents~\cite{zhou2023malib}. In section~\ref{sec:pytag} we highlighted some of the challenges TTGs present in general for MARL approaches. We believe that PyTAG, along with TAG, have the generality required to study MARL through TTGs. 

In MARL, the agent's performance is measured against some opponent. Here, we benchmarked our trained agents against TAG baselines. Our results show that even in this setting, training an RL agent is difficult in most games. ``Stratego" is a challenging strategy game, where the RL agent found a good policy against the baseline agents early on, which is likely due to the baseline agents struggling to play well due to the game length. These tests could be done against stronger opponents or with increased budgets for the baseline agents (i.e: MCTS with more budget), but then it becomes difficult to measure how good agents are, as all the comparisons are relative. Therefore, measuring match-making ranking such as Trueskill~\cite{herbrich2006trueskill} is a good direction for future work. 

Tabletop games are often released with limited opportunity to correct mistakes once they are produced. AI methods could provide a tool to test such games by allowing them to find exploits in the games which could be tracked using various game-specific metrics, such as which cards were used when the player won or key events that happened in the game. In addition to testing, novel tabletop games often come with expansions which may alter the game in a significant way by adding new rules, or just by adding a new game board or new cards. PyTAG may provide a new framework for generative AI methods, to create more content for games while evaluating them using game-playing agents. 

\section{Conclusion}
This work presented PyTAG, a collection of games with a Reinforcement Learning interface to interact with a large collection of TTGs implemented in the TAG framework. In this work, we explored the challenges and opportunities that Tabletop games present for training RL agents. Compared to previous works, PyTAG is the first framework to provide a collection of various modern board games, card games and potentially role-playing games in the future. 
Our aim with PyTAG is to keep the entry barrier low for the community to do research in the framework. All the code used in this work is publicly available on GitHub. Setting up PPO with self-play presented a couple of technical challenges, mainly due to the changing turn order while maintaining high training speeds with the vectorised environments. As future work, we believe that adding explicit memory to the RL agents could boost their performance on some of the games presented in our experiments. We presented results on a set of games which we believe characterised some of the challenges, but not all. As future work, exploring games that require more cooperation or mixed aspects would be a valuable extension. TAG contains a set of fully cooperative games, such as ``Hanabi" and ``Pandemic", and more complicated strategy games such as ``Settlers of Catan", ``Battlelore and ``Terraforming Mars". 

The experiments presented in this work focused on training the PPO algorithm with self-play and evaluated against the baseline agents implemented in the TAG framework. With these experiments, our objective was not to develop the state-of-the-art agent on any of the games in particular but to explore the challenges these games present. As future work, the self-play setting could be improved by introducing a weighted sampling based on recency instead of using a hyperparameter to bias the sampling towards choosing the latest checkpoint with higher probability. As self-play may not fully capture the strategic depth of all the games in TAG, exploring a league-based system~\cite{alphastar} could be an interesting direction. 

In our evaluations, we used some simple baseline agents and MCTS to evaluate against, but the latter did not use any of the extensions that were proposed to improve performance for tabletop games. As future work with stronger RL agents, the evaluations could be also improved by using stronger baselines tuned specifically for the games. Looking at the win rates and scores in these multi-player games alone do not give enough insight to understand the RL agent's capabilities, as all the results are relative to the opponents' skill. As future work, more game-specific metrics could be used to measure the agent's skill level. Additionally, a match-making ranking, such as TrueSkill~\cite{herbrich2006trueskill} could be calculated to make better comparisons among agents.

\section*{Acknowledgment}
Work supported by the EPSRC IGGI CDT (EP/S022325/1). For the purpose of open access, the author(s) has applied a Creative Commons Attribution (CC BY) license to any Accepted Manuscript version arising.

\ifCLASSOPTIONcaptionsoff
  \newpage
\fi

\bibliographystyle{IEEEtran}
\bibliography{main}

\end{document}